\documentclass[letterpaper, 10 pt, conference]{ieeeconf}  
\usepackage{graphicx}
\usepackage{float}
\graphicspath{ {./} }
\usepackage{booktabs}
\usepackage[dvipsnames]{xcolor}
\usepackage{tikz}

\IEEEoverridecommandlockouts                              

\begin{document}
\title{\LARGE \bf
Video Colorization using CNNs and Keyframes extraction: An application in saving bandwidth
}
\author{Ankur Singh$^{1}$  Anurag Chanani$^{2}$  Harish Karnick$^{3}$
}

\maketitle
\thispagestyle{empty}
\pagestyle{empty}


\begin{abstract}

\textit{In this paper, we tackle the problem of colorization of grayscale videos to reduce bandwidth usage. For this task, we use some colored keyframes as reference images from the colored version of the grayscale video. We propose a model that extracts keyframes from a colored video and trains a Convolutional Neural Network from scratch on these colored frames. Through the extracted keyframes we get a good knowledge of the colors that have been used in the video which helps us in colorizing the grayscale version of the video efficiently. An application of the technique that we propose in this paper, is in saving bandwidth while sending raw colored videos that haven't gone through any compression. A raw colored video takes up around three times more memory size than it’s grayscale version. We can exploit this fact and send a grayscale video along with our trained model instead of a colored video. 
Later on, in this paper we show how this technique can help to save bandwidth usage to upto three times while transmitting raw colored videos.\newline
}
\end{abstract}

\section{INTRODUCTION}
Learning based colorization algorithms for grayscale videos and images have
been the subject of extensive research in the areas of computer vision
and machine learning. Apart from being
alluring from an artificial intelligence point of view, such potential has vast practical implementations
starting from video restoration to image improvement for
enhanced understanding.
Colorizing a grayscale image can be hugely beneficial, since grayscale images contain very less information thus adding color can add a lot of information about the semantics. \par
Another motivation for video colorization that we propose, is it's capacity to save data while transmitting a video. A raw colored video takes upto three times more memory than it's grayscale version. Hence sending a grayscale video instead of a colored one while streaming and then colorizing it on the receiver's end can help save data and in turn the bandwidth. 
In this paper, we propose a convolutional neural network model that is trained on the keyframes of a raw colored video. This model is transmitted along with the grayscale version of the colored video and on the receiver's end this model colorizes the grayscale video. 
Apart from our convolutional neural network model we also propose a keyframe extraction method that extracts keyframes from a video by comparing colored histograms of all the frames in that video.\newline
Also, in general in image and video colorization a given grayscale image can have varying colored outputs when tested with different colorization models. For eg. a grayscale image of a ball can have different colored outputs from different colorizing models. Some models may output a green colored ball while some may output a blue colored ball. This might differ from the actual color of the ball. Hence, in this paper we also tackle this problem by using few colored keyframes of the video to colorize the grayscale video. Having a sense of the colors that have been used in the video will help a great deal in predicting the actual colors of the rest of the frames of the video.

Hence our work serves two purpose:
\begin{itemize}
  \item Saving bandwidth while transmitting a video by sending grayscale version of a raw colored video along with a CNN model trained on the keyframes of the video and then colorizing the grayscale video on the receiver's end as shown in Fig. 1.
  \begin{figure}[t]
\centering
\includegraphics[width=\linewidth, height=10cm]{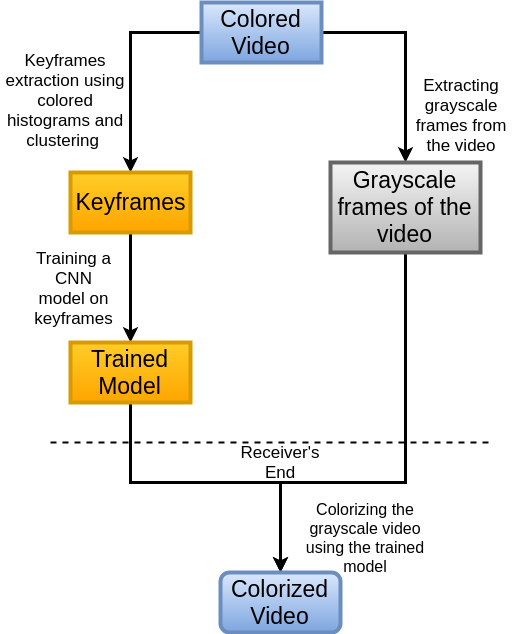}
\caption{Sending grayscale version of a raw colored video  alongwith a convolutional neural network model trained  on keyframes of the video and then colorizing the grayscale video on the receiver’s end.}
\end{figure}
  
  \item Tackling the problem of different colored outputs of the same image(shown in Fig. 2) by colorizing a video using few keyframes of the video.
\end{itemize}
\begin{figure}[H]

\includegraphics[width=\linewidth, height=3.5cm]{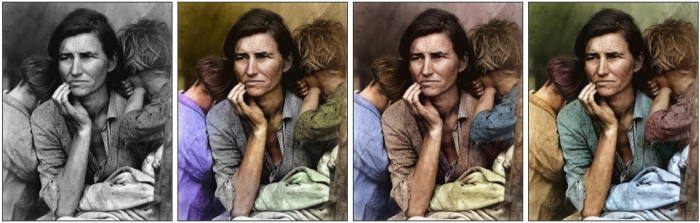}
\caption{A grayscale image can have multiple colored output image from Alexei Afros homepage}
\end{figure}


\section{Previous Work}
The start of automated image colorization can be dated back to 2002, when Welsh et. al. \cite{welsh2002transferring} presented an approach which could colorize a grayscale image by transferring colors from a related reference image.
Our work on video colorization is inspired by Baldassarre et. al. \cite{baldassarre2017deep} system on automatically colorizing images. Apart from the convolutional neural network that we have used, they have also employed Inception Resnet \cite{szegedy2017inception} as a high level feature extractor which provides information about the image contents that helps in their colorization. Their network consists of four main components: an encoder, a feature extractor, a fusion layer and a decoder. The encoding
and the feature extraction parts obtain mid and high-level features, respectively,
which are then merged in the fusion layer. Finally, the decoder uses
these features to estimate the output. Iizuka et. al. \cite{iizuka2016let} and Larsson et. al. \cite{larsson2016learning} have developed similar models.
Zhang et. al. \cite{zhang2016colorful} use a classification loss in their architecture unlike the regression loss that we have used. 
\par
The work on Keyframes extraction is inspired from Zhuang et. al. \cite{zhuang1998adaptive} work on color histograms. In a color histogram, a 1D array contains the total pixels that belong to a particular color in the image. All the images are resized to the same shape before their histograms are taken so that they have equal number of pixels. To discretize the space, images are represented in RGB colorspace using some important bits for every color component. 
\par
The main purpose for which we've employed color histograms in keyframes extraction is that they are very easy to compute and show striking properties despite their simplicity. They are often used for content based image retrieval. They are also highly invariant to the translation and rotation of objects in the image since they do not relate spatial information with the pixels of the colors.

\section{Proposed Method}

We introduce a two step process for our approach of colorizing grayscale videos using keyframes extraction.\par
The first step deals with the extraction of keyframes of the video. The second step involves training a Convolutional Neural Network on these keyframes and colorizing the rest of the video using the trained model.
\subsection{Keyframes Extraction} We extract keyframes of a video by comparing colored histograms of all frames with a sample image. In our experiments we have taken the sample image to be a black image(all pixels equal to zero). \par We extract a 3D RGB color histogram with 8 bins per channel for all the frames. This yields a 512-dimensional feature vector for a frame once flattened. For comparing two histograms we use the Hellinger distance which is used to measure the “overlap” between the two histograms. 

Formally, let H be the 512 dimensional colored histogram of our sample image. Let h\textsubscript{i} be the 512 dimensional colored histogram of the i\textsuperscript{th} frame. We calculate the Hellinger distance \textit{d(H, h\textsubscript{i})} between H and h\textsubscript{i} by:

\vspace{8mm}

\textit{d(H, h\textsubscript{i})} = $\sqrt{1-\frac{1}{\sqrt{\overline{H}\,\overline{h\textsubscript{i}}} N\textsuperscript{2}}\sum_{j=0}^{511} \sqrt{H[j]h\textsubscript{i}[j]}}$\\

\vspace{5mm}
$N$ = total number of bins of the histogram, $\overline{x} = \frac{1}{N}\sum_{j}x[j]$ \newline

Additionally, we multiply the hellinger distance by a factor of 10,000 to ease out calculations that follow this step. 

Once, we have the distances for all the frames against our sample image we use mean shift clustering \cite{cheng1995mean} to cluster frames whose distances from the random image are close to each other. The mean shift algorithm is a non parametric clustering technique that does not need initial 
information about the number of clusters. This property is essential in our problem since we don't have any prior knowledge about the number of clusters present in a particular video. Result of clustering on a 1 minute video is shown in Fig. 3.
\begin{figure}[t]

\includegraphics[width=\linewidth, height=5cm]{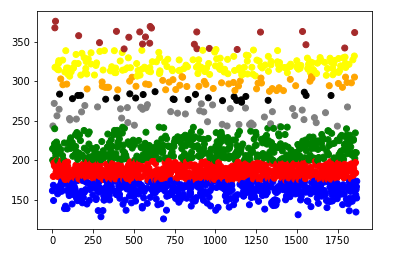}
\caption{Result of Meanshift algorithm applied on the frames of the video, Clusters have been represented in different colors. X axis represents indices of the frames. Y axis denotes the Hellinger distance of a frame from the sample image.}
\end{figure}

After we have the clusters we can choose every x\textsuperscript{th} frame from the cluster depending upon the number of frames we want. We have found emperically that x equal to 30 does a good job.\newline

\begin{figure*}[t]
  \includegraphics[width=\textwidth,height=4cm]{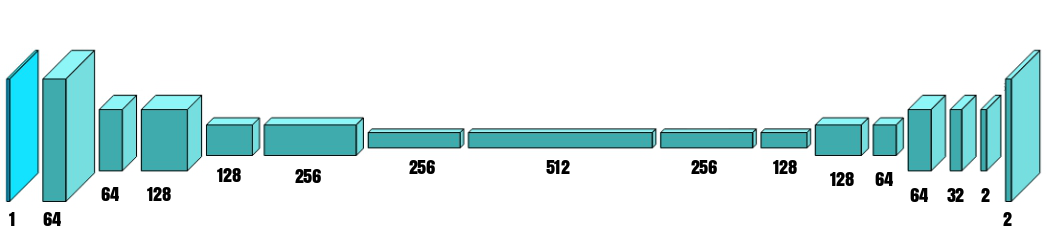}
  \caption{Architecture of the network: 12 convolutional and 3 upsampling layers have been used. In the 2nd 4th and the 6th convolutional layer a stride of 2 has been applied. A final image with dimensions H x W x 2 is obtained. The 2 output channels are merged with the L component to get the final colored image. }
\end{figure*}
\subsection{Training a Convolutional Neural Network}
For the training part, we consider images in the CIELab color space. Here L stands for lightness, a stands for the green red color spectra and b stands for the blue yellow color spectra. A CIELab encoded image has one layer for grayscale, and it packs three color layers into two. This means that the original grayscale image can be used in our final prediction. Also, we only have two channels to predict. Starting
from the L component X\textsubscript{L}, the purpose of our model is to estimate the remaining two components X\textsubscript{a} and X\textsubscript{b}.\newline

\textbf{Preprocessing}\newline
The pixel values of all three image
components namely L, a and b are centered and scaled to get values within the [{-1}, 1] range. All images are converted from RGB color space to CIELab color space to feed them into our model.  \newline

\textbf{Architecture}\newline
The architecture of our model is inspired from [1]. Given the L component of an image, our model estimates it's a and b components and combines them with the L component to get the final colored image.
We have used 12 convolutional layers with 3 x 3 kernels and 3 upsampling layers as shown in Fig. 4. In the second, fourth and the sixth convolutional layer, a stride of two is applied which halves the dimension of their output, resulting in less number of computations \cite{DBLP:journals/corr/SpringenbergDBR14}. We have made use of padding to preserve the layer’s input dimension. Upsampling has been performed so that the height and width of the output are twice that of the input. This model applies a number of convolutional and upsampling layers in order to output a final image with dimensions H x W x 2. The 2 output channels are a and b. These are merged with the L component to get the colored image. \newline

\textbf{Training}\newline
We obtain the optimal parameters of the model by minimizing a function which is defined over the predicted output of our network and the target output. In order to quantify the model loss, we employ the mean squared loss between the estimated pixel
colors in a, b space and their real value. While training, we back propagate this loss to update the model parameters  using Adam Optimizer \cite{kingma2014adam} with a learning rate of 0.001. During
training, we impose a fixed input image size to allow for batch processing.

\begin{table}[H]
\centering
\tiny
\resizebox{5.5cm}{3cm}{
\begin{tabular}{|c|c|c|}
\hline
\textbf{Layer} & \textbf{Kernels} & \textbf{Stride} \\ \hline
Convolution           & (64, 3, 3)       & (1, 1)          \\ \hline
Convolution           & (64, 3, 3)       & (2, 2)          \\ \hline
Convolution           & (128, 3, 3)      & (1, 1)          \\ \hline
Convolution           & (128, 3, 3)      & (2, 2)          \\ \hline
Convolution           & (256, 3, 3)      & (1, 1)          \\ \hline
Convolution           & (256, 3, 3)      & (2, 2)          \\ \hline
Convolution           & (512, 3, 3)      & (1, 1)          \\ \hline
Convolution           & (256, 3, 3)      & (1, 1)          \\ \hline
Convolution           & (128, 3, 3)      & (1, 1)          \\ \hline
Upsampling         & -                & -               \\ \hline
Convolution           & (64, 3, 3)       & (1, 1)          \\ \hline
Upsampling         & -                & -               \\ \hline
Convolution           & (32, 3, 3)       & (1, 1)          \\ \hline
Convolution           & (2, 3, 3)        & (1, 1)          \\ \hline
Upsampling         & -                & -               \\ \hline
\end{tabular}}
\caption{Architecture of the network}
\end{table}

\section{Experiments and Results}

We tested our model on a number of videos. For a 256x256 24 bit 15 minute uncompressed colored video that has a size of around 5 GB, we could save a bandwidth of around 3.30 GB as our trained model had a size of only 30MB. Also, it took us only around 6 minutes for the whole process starting from keyframes extraction to training a model and finally obtaining the colored output video for a 256x256 24 bit 15 minute video on NVIDIA GeForce GTX 1080. 
\newline
Since, our main aim was to reduce the model size so that we could save as much bandwidth as possible we kept our CNN model simple, without hampering the quality of the colored video that we output. The results turned out to be quite good for most of the videos. However, the videos in which there were drastic changes from one shot to another, our network was not able to produce that good results. We observed that although some results
were quite good, some generated pictures tend to be low saturated, with the network producing a grayish color where the original was brighter. 
\begin{table*}[t]
\centering       
        \begin{tabular}{ccccc}
           \toprule
            Grayscale & Ground truth & Zhang et.al & Ours\\
            \midrule
            \includegraphics[width=3.5cm, height=2.2cm]{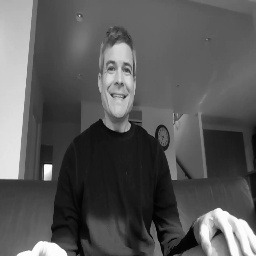} & \includegraphics[width=3.5cm, height=2.2cm]{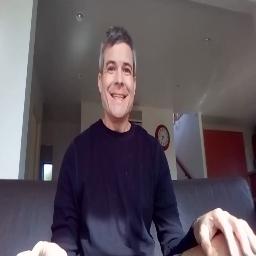} &
            \includegraphics[width=3.5cm, height=2.2cm]{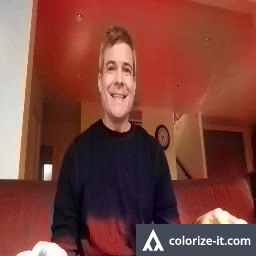} &
            \includegraphics[width=3.5cm, height=2.2cm]{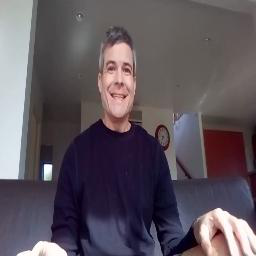} \cite{Video1}\\
            
            \includegraphics[width=3.5cm, height=2.2cm]{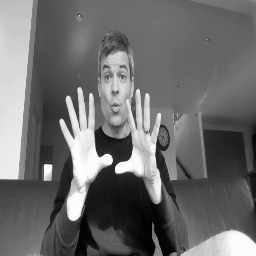} & \includegraphics[width=3.5cm, height=2.2cm]{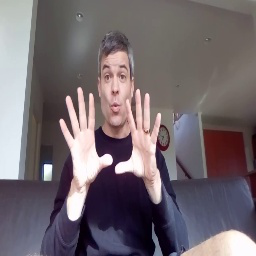} &
            \includegraphics[width=3.5cm, height=2.2cm]{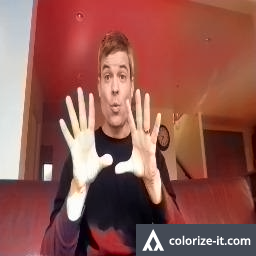} &
            \includegraphics[width=3.5cm, height=2.2cm]{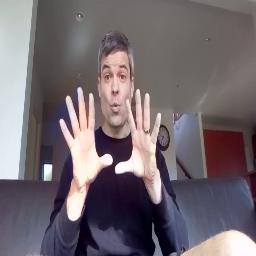} \cite{Video1}\\
            
            \includegraphics[width=3.5cm, height=2.2cm]{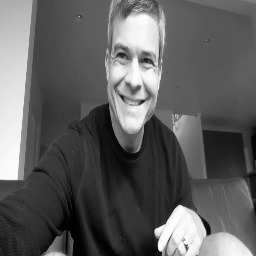} & \includegraphics[width=3.5cm, height=2.2cm]{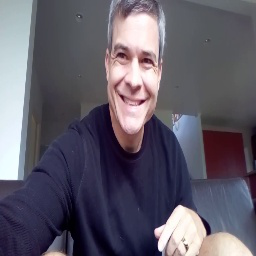} &
            \includegraphics[width=3.5cm, height=2.2cm]{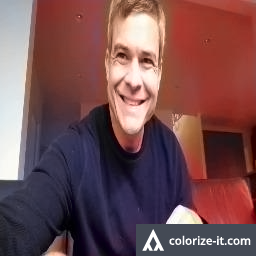} &
            \includegraphics[width=3.5cm, height=2.2cm]{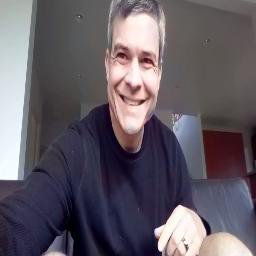} \cite{Video1}\\
            
            \includegraphics[width=3.5cm, height=2.2cm]{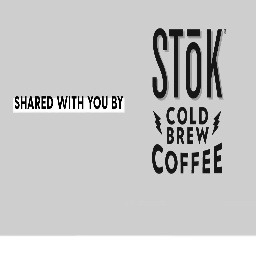} & \includegraphics[width=3.5cm, height=2.2cm]{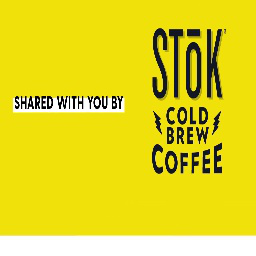} &
            \includegraphics[width=3.5cm, height=2.2cm]{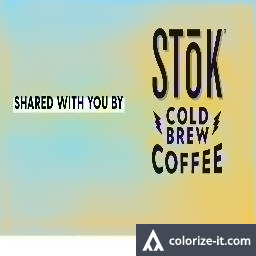} &
            \includegraphics[width=3.5cm, height=2.2cm]{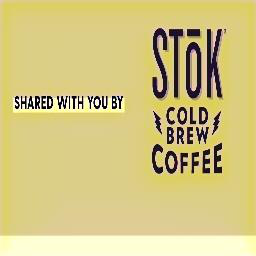} \cite{Video2}\\
            
            \includegraphics[width=3.5cm, height=2.2cm]{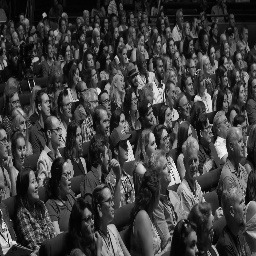} & \includegraphics[width=3.5cm, height=2.2cm]{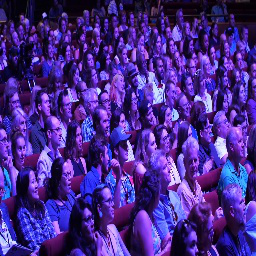} &
            \includegraphics[width=3.5cm, height=2.2cm]{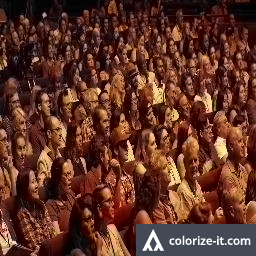} &
            \includegraphics[width=3.5cm, height=2.2cm]{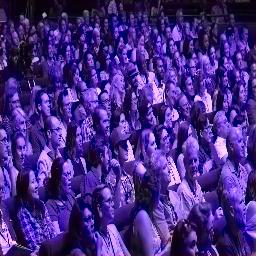} \cite{Video2}\\
            
            \includegraphics[width=3.5cm, height=2.2cm]{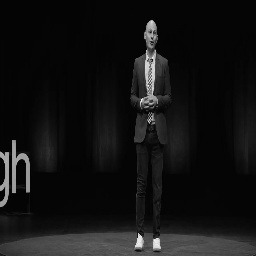} & \includegraphics[width=3.5cm, height=2.2cm]{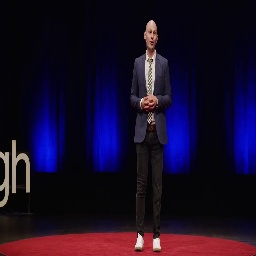} &
            \includegraphics[width=3.5cm, height=2.2cm]{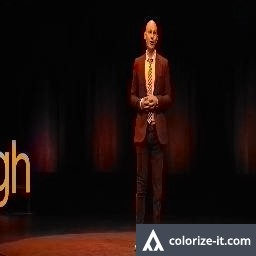} &
            \includegraphics[width=3.5cm, height=2.2cm]{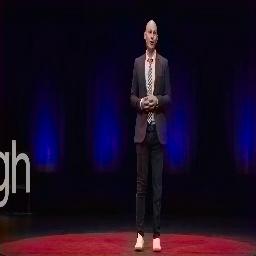} \cite{Video2}\\
            
            \includegraphics[width=3.5cm, height=2.2cm]{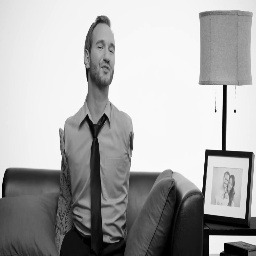} & \includegraphics[width=3.5cm, height=2.2cm]{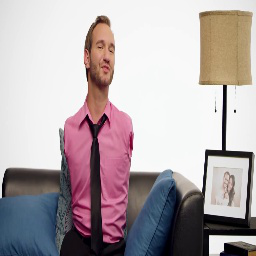} &
            \includegraphics[width=3.5cm, height=2.2cm]{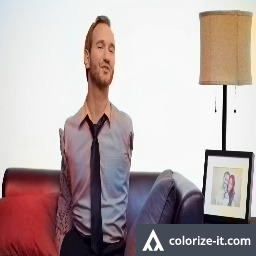} &
            \includegraphics[width=3.5cm, height=2.2cm]{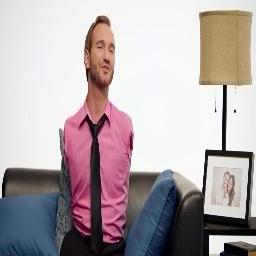} \cite{Video3}\\
            
            \includegraphics[width=3.5cm, height=2.2cm]{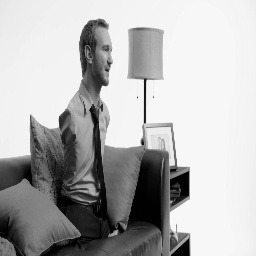} & \includegraphics[width=3.5cm, height=2.2cm]{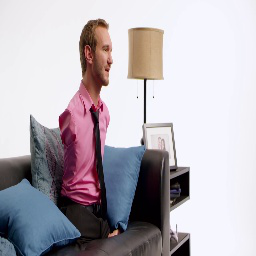} &
            \includegraphics[width=3.5cm, height=2.2cm]{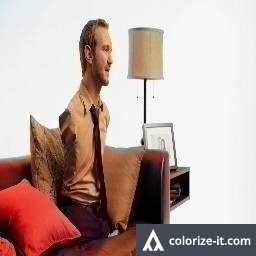} &
            \includegraphics[width=3.5cm, height=2.2cm]{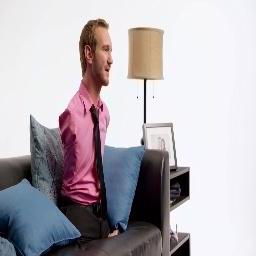} \cite{Video3}\\
            
            \includegraphics[width=3.5cm, height=2.2cm]{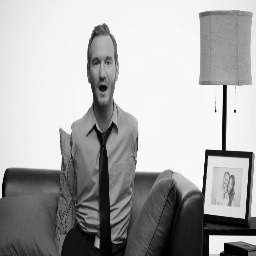} & \includegraphics[width=3.5cm, height=2.2cm]{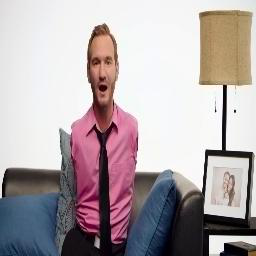} &
            \includegraphics[width=3.5cm, height=2.2cm]{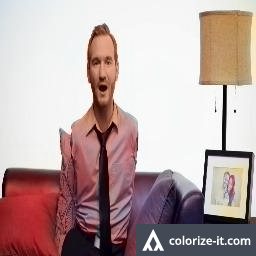} &
            \includegraphics[width=3.5cm, height=2.2cm]{1319} \cite{Video3}\\
            \bottomrule
            
            \bottomrule
        \end{tabular}
        \caption{Results}
        \label{tbl:table_of_figures}
        
    \end{table*}

\begin{table*}[t]
\centering

\resizebox{16cm}{1.5cm}{
\begin{tabular}{|l|l|l|l|l|l|}
\hline
\textbf{\begin{tabular}[c]{@{}l@{}}Input Video \\ frame size\end{tabular}} & \textbf{\begin{tabular}[c]{@{}l@{}}Input \\ Video duration\end{tabular}} & \textbf{\begin{tabular}[c]{@{}l@{}}Time taken \\ to output \\ colored video\end{tabular}} & \textbf{\begin{tabular}[c]{@{}l@{}}Size of \\ model\end{tabular}} & \textbf{\begin{tabular}[c]{@{}l@{}}Bandwidth\\ saved\end{tabular}} & \textbf{\begin{tabular}[c]{@{}l@{}}Percentage \\ Bandwidth\\ saved\end{tabular}} \\ \hline
256X256 (24 bit)                                                           & 1 minute                                                                 & $\sim$4 minutes                                                                                  & 30MB                                                              & 195MB &57.78\%                                                           \\ \hline
256X256 (24 bit)                                                           & 15 minutes                                                               & $\sim$6 minutes                                                                           & 30MB                                                              & 3345MB & 66.07\%                                                         \\ \hline
720X1280 (24 bit)                                                          & 15 minutes                                                               & $\sim$6 hours                                                                             & 45MB                                                              & 46.3GB & 66.6\%                                                         \\ \hline

\end{tabular}}
{
\caption{Results of bandwidth saved in various videos through our approach}}
\end{table*}

Table II shows some of our results. The varying colored outputs of similar grayscale images has been clearly shown in the output of Zhang et. al. model. The ground truth of the last and the second last images have similar colors. However, Zhang et. al. outputs an image with a pinkish shade in one case and an image with a reddish tint in the other. We easily handle this anomaly since we already have a knowledge of the colors that have been used in the video, that we extract through the keyframes.
\newline
In Table III we have shown the bandwidth that is saved through our approach. The small size of the trained model helps in accomplishing our task to a great extent.


\section{Conclusion and Future Work}
In this paper, we devised a new approach to save bandwidth to upto three times while transferring colored videos without losing data or hampering the quality of the video. Usual compression algorithms are lossy, hence lose data while compressing videos. Lossy compressions are irreversible that use inaccurate estimations and discard some data to present the content. They are performed to decrease the size of data for storing, handling, and transmitting content. However the approach that we propose isn't irreversible in the sense that the quality of the output video is not hampered.
We also tackled the problem of varying colored outputs of a single grayscale frame of a video when tested with different colorization models, by using some colored keyframes of the video as reference images. Having a knowledge of the colors that have been used in a video will help in colorizing the rest of the frames of the video.

Our future work will focus on reducing the time taken to output the colored video through our model without trading off with the quality. We will also work on colorizing videos where drastic changes occur from one shot to another, in a better way.

\addtolength{\textheight}{-12.5cm}   



\bibliographystyle{abbrv}
\bibliography{ref}

\begin{thebibliography}{10}

\bibitem{baldassarre2017deep}
F.~Baldassarre, D.~G. Mor{\'\i}n, and L.~Rod{\'e}s-Guirao.
\newblock Deep koalarization: Image colorization using cnns and
  inception-resnet-v2.
\newblock {\em arXiv preprint arXiv:1712.03400}, 2017.

\bibitem{cheng1995mean}
Y.~Cheng.
\newblock Mean shift, mode seeking, and clustering.
\newblock {\em IEEE transactions on pattern analysis and machine intelligence},
  17(8):790--799, 1995.

\bibitem{Video1}
https://www.youtube.com/watch?v=EWg52nptETc.

\bibitem{Video2}
https://www.youtube.com/watch?v=EWg52nptETc.

\bibitem{Video3}
https://www.youtube.com/watch?v=i5Q02YX2VTw.

\bibitem{iizuka2016let}
S.~Iizuka, E.~Simo-Serra, and H.~Ishikawa.
\newblock Let there be color!: joint end-to-end learning of global and local
  image priors for automatic image colorization with simultaneous
  classification.
\newblock {\em ACM Transactions on Graphics (TOG)}, 35(4):110, 2016.

\bibitem{kingma2014adam}
D.~P. Kingma and J.~Ba.
\newblock Adam: A method for stochastic optimization.
\newblock {\em arXiv preprint arXiv:1412.6980}, 2014.

\bibitem{larsson2016learning}
G.~Larsson, M.~Maire, and G.~Shakhnarovich.
\newblock Learning representations for automatic colorization.
\newblock In {\em European Conference on Computer Vision}, pages 577--593.
  Springer, 2016.

\bibitem{DBLP:journals/corr/SpringenbergDBR14}
J.~T. Springenberg, A.~Dosovitskiy, T.~Brox, and M.~A. Riedmiller.
\newblock Striving for simplicity: The all convolutional net.
\newblock {\em CoRR}, abs/1412.6806, 2014.

\bibitem{szegedy2017inception}
C.~Szegedy, S.~Ioffe, V.~Vanhoucke, and A.~A. Alemi.
\newblock Inception-v4, inception-resnet and the impact of residual connections
  on learning.
\newblock In {\em AAAI}, volume~4, page~12, 2017.

\bibitem{welsh2002transferring}
T.~Welsh, M.~Ashikhmin, and K.~Mueller.
\newblock Transferring color to greyscale images.
\newblock In {\em ACM Transactions on Graphics (TOG)}, volume~21, pages
  277--280. ACM, 2002.

\bibitem{zhang2016colorful}
R.~Zhang, P.~Isola, and A.~A. Efros.
\newblock Colorful image colorization.
\newblock In {\em European Conference on Computer Vision}, pages 649--666.
  Springer, 2016.

\bibitem{zhuang1998adaptive}
Y.~Zhuang, Y.~Rui, T.~S. Huang, and S.~Mehrotra.
\newblock Adaptive key frame extraction using unsupervised clustering.
\newblock In {\em Image Processing, 1998. ICIP 98. Proceedings. 1998
  International Conference on}, volume~1, pages 866--870. IEEE, 1998.

\end{thebibliography}

\end{document}